\documentclass[11pt]{article}

\usepackage[final]{acl}

\usepackage{times}
\usepackage{latexsym}

\usepackage[T1]{fontenc}

\usepackage[utf8]{inputenc}

\usepackage{microtype}

\usepackage{subcaption}

\usepackage{inconsolata}

\usepackage{graphicx}

\usepackage{booktabs}
\usepackage{multirow}
\usepackage{fancyvrb}
\usepackage{fvextra}
\title{Closing Cost-Quality Gap in Document VLMs: Difficulty-Aware Data Curation and Quality-Adjusted Deployment Economics}

\author{
    \textbf{Maksim Evdokimov}, \textbf{Matvey Ivanov}, \textbf{Dmitrii Tsiupin}, \textbf{Olga Tsymboi}, \\ 
    \textbf{Anatolii Potapov}, 
    \textbf{Aleksandr Ivanov} \\ 
    T-Tech \\
 \small{
   \textbf{Correspondence:} \href{mailto:m.v.evdokimov1@gmail.com}{m.v.evdokimov1@gmail.com}
 }
}

\begin{document}
\maketitle
\begin{abstract}
Extracting structured fields from hundreds of millions of documents annually remains costly in regulated industries: bespoke OCR cascades cover only a fraction of workflows, privacy rules preclude external models, and existing open-source VLMs that clear quality thresholds cost more to serve than human annotation. We present a deployed document-understanding system built on a Mixture-of-Experts VLM (35B total, 3B active), fine-tuned on in-house production data mixed with open-domain documents curated by a Difficulty-Aware pipeline for layout diversity, fact-extractability, and cross-model consistency. Fitting on a single H100 and serving heterogeneous workflows via prompting, the model leads all deployable (non-reasoning) baselines up to an order of magnitude larger. A quality-adjusted cost analysis, with confirmation and correction costs calibrated from production telemetry, shows it reduces expected costs by over 80\% against the human baseline and by more than 50\% against the best competing open-source model, while larger baselines remain economically unviable.
\end{abstract}

\section{Introduction}
Vision--language models (VLMs) are improving quickly at documents understanding \citep{bai2025qwen25vl, qwen3vl2025, qwen35blog}, but using them in regulated business settings is still hard because of high serving costs, strict latency constraints and privacy rules. We tackle these problems at a large document processing platform which processes hundreds of millions of Cyrillic-language documents each year, including court judgments, invoices, tax certificates, and receipts. Human workers extract structured fields, sort documents by type, and check visual elements. Four barriers block automation. Documents contain personal data, which rules out cloud-hosted models. Open-source VLMs under 10B fall short of production quality, while larger models that clear quality thresholds are prohibitively expensive to serve at document-flow scale. The standard pipeline of OCR plus specialized models \citep{yoon2024lofi, wang2025hybridocr} is costly to maintain and outgrows engineering capacity. Strict per-document time limits make reasoning-mode inference too slow for production.

We introduce a domain-adapted Mixture-of-Experts (MoE) VLM with 35B total and 3B active parameters that replaces separate workflow pipelines with a single prompt-driven model on one H100. Training mixes internal production data, where labeled fields come as a byproduct of daily work, with open-domain Common Crawl PDFs. The internal data matches production needs but saturates quickly; Common Crawl adds broad variety, yet samples generated without filtering are mostly too easy and lower quality. Our Difficulty-Aware Data Curation (DADC) pipeline solves this by keeping only hard, information-rich samples.

The final model averages 0.814 across all evaluation groups, 8.4 points above its base checkpoint and 5.3 above the strongest reasoning-mode baseline, which is ten times larger. On non-fine-tuned groups alone, it leads the base by 3.4 points and the reasoning baseline by 2.4. It is deployed and handles production traffic within latency constraints. A cost analysis adjusted for quality shows that large baselines need so many GPUs that they cost more than the human work they replace, while our model beats the strongest deployable baseline by 53\%.

Our contribution is a unified cost-aware methodology for building a deployable document VLM under strict hardware and privacy constraints. The approach begins with model selection guided by per-document economics, then scales quality through a DADC pipeline that converts commodity PDFs into high-signal supervision. The resulting model beats baselines on in-house benchmarks up to $4\times$ larger in active parameters and $10\times$ in total paramaters while needing $8\times$ fewer GPUs, and a quality-adjusted cost framework, calibrated from production telemetry, ties per-field accuracy directly to deployment economics. The methodology is workflow-independent and closed-form, with adaptation to a given workflow requiring only recalibration of the coefficients (workload distribution, GPU utilization, routing policy, per-action costs). We instantiate it with coefficients estimated from our production traffic and report the resulting post-deployment costs.

\section{Related Work}

General-purpose VLMs have advanced rapidly on document understanding: the Qwen family~\citep{bai2025qwen25vl,qwen3vl2025} provides the multilingual foundation we build upon, and frontier VLMs~\citep{seed15vl2025,glm41v2025,ernie45,qwen35blog} confirm that scaling yields consistent benchmark gains, though 70B--400B deployment requires multi-GPU serving with prohibitive per-document cost. MoE architectures offer a more efficient alternative: DeepSeek-VL2~\citep{wu2024deepseekvl2} matches dense models at 4.5B active parameters, MoE-LLaVA~\citep{lin2024moellava} established MoE-Tuning for initializing sparse experts from dense FFNs, and the DeepSeek-OCR line~\citep{wei2025deepseekocr,wei2026deepseekocr2} pushes this further with a 3B MoE decoder achieving strong OCR accuracy under tight token budgets via optical context compression. olmOCR~\citep{poznanski2025olmocr} distills GPT-4o into a 7B VLM using native PDF text layers as supervision, conceptually related to our text-to-image substitution, and olmOCR~2~\citep{poznanski2025olmocr2} adds RL with unit-test rewards, HunyuanOCR~\citep{hunyuanocr2025}, MinerU2.5-Pro~\citep{miner25pro}, and Qianfan-OCR~\citep{qianfanocr2026} explore further points in the quality-efficiency space. However, domain adaptation of VLMs for regulated settings where privacy precludes external APIs, sub-10B models miss production thresholds, and a single  model in prompting must serve a growing number of heterogeneous workflows remains the open challenge we address.

Several recent systems deploy VLMs in production: VENUS~\citep{zhao2025venus} consolidates per-policy video classifiers into a single instruction-tuned VLM with synthetic annotation paralleling our per-workflow unification. IPL~\citep{chen2024ipl} adapts a multimodal LLM for C2C product listing reporting 72\% user adoption, and PlanGPT-VL~\citep{zhu2025plangptvl} fine-tunes a 7B VLM for urban planning maps, rivaling 72B models via structured data synthesis.
For document extraction,~\citet{yoon2024lofi} and \citet{wang2025hybridocr} build cascaded OCR--LLM pipelines for insurance, tax, and enterprise settings; our system replaces such cascades with end-to-end extraction from rendered pages via a single VLM.
On data curation,~\citet{dong2025sailvl} shows that quality-filtered, progressively complex data outperforms raw scaling, similarly our DADC pipeline specializes to the document domain, where corrupt text layers, skewed layout distributions, and generator hallucinations require domain-specific filtering stages.

\citet{chen2023frugalgpt} cascades queries across LLMs under pay-per-token pricing. Confidence-based routing to human review is common in commercial platforms, but asymmetric confirmation-versus-correction costs and per-field economics against a manual baseline remain largely unformalized, the gap our framework addresses.

\section{Framework}

Building a production document VLM requires both scalable, production-representative supervision and tight hardware and cost budgets at document-flow scale. We address both through SFT on a mixture of in-house and open-domain data across four canonical tasks: schema-based field extraction with absent-field handling, closed-set document classification, visual element validation (checkboxes, stamps, signatures), and free-form VQA/OCR, all served through a single instruction-following interface.

\subsection{Internal Data}

Documents processed at inference time flow through operational workflows in which field values are captured as part of routine business operations, yielding naturally aligned supervision whose semantics and taxonomies match the evaluation distribution by construction, with no additional annotation effort. The internal corpus is drawn from two such workflows: a judicial pipeline (60K court PDFs, 20 fields, ~8 per document) and an invoice processing pipeline (150K instances spanning PDFs, scans, screenshots, and photos, ~4--5 fields each). The corpus is predominantly Russian. Each training sample pairs an instruction with a ground-truth response. The instruction comprises a task description covering field semantics, layout cues, and disambiguation rules, together with an output schema encoding per-field formatting constraints and absent-value handling. To mitigate overfitting to fixed templates, prompt components are sampled from sets of semantically equivalent alternatives and key-value pairs are randomly permuted. All annotations undergo quality control with overlapping annotation (3 of 5 annotators per item). The complete prompt template is given in Appendix~\ref{app:internal}.

\subsection{Synthetic Data Pipeline}
\label{sec:dadc}

Internal data offers limited diversity: layouts follow operational templates and field taxonomies are fixed by business processes, so both saturate quickly during training (Section~\ref{sec:internal_scaling}). We therefore construct a large-scale open-domain synthetic dataset from 300K Common Crawl~(CC) PDFs filtered to Russian, Belarusian, Ukrainian, and Kazakh via GlotLID~\citep{kargaran2023glotlid}, using a Difficulty-Aware Data Curation (DADC) pipeline. The key insight is that naively generated samples are largely solvable by the base model and add negligible signal: unfiltered open data in fact degrades the in-house--only baseline (Appendix~\ref{app:dadc-ablation}), so DADC filters aggressively to retain only challenging, information-rich examples.

\paragraph{Text layer validation.} PDF embedded text is frequently corrupt or misaligned --- silently poisoning generated data via malformed encodings or transparent OCR layers from digitization pipelines. We run an internal OCR engine on rendered pages and perform a word-wise consistency check against the embedded text layer; pages below the alignment threshold are discarded. The stage retains 40\% of input pages. The complete procedure is in Appendix~\ref{app:text-layer}.

\paragraph{Document Sampling.} Validated pages are dominated by continuous prose, a poor match for our deployment, where fields must be extracted from forms, tables, and dense structured layouts. We rebalance the pool via a two-stage filter built on VLM judge (Qwen2.5-VL-72B-Instruct, \cite{bai2025qwen25vl}). \emph{Stage 1 -- taxonomic classification} labels pages from a fixed taxonomy (Tabular, Charts, Logic Diagrams, Schematic, Infographic, Plain Text); any non-plain text label admits the page. \emph{Stage 2 -- fact-extractability} re-scores remaining plain text pages ($\sim$ 70\% of the pool) on visual structure and QA potential, admitting those scoring greater than given threshold -- recovering fact-dense but structurally plain documents the structural filter alone discards. Details are in Appendix~\ref{app:document-sampling}.

\paragraph{Text-to-image substitution.} Query–response pairs are generated from the embedded PDF text layer, then the text representation is replaced by document pages rendered via \texttt{fitz} (PyMuPDF) — grounding supervision in visual understanding while leveraging the reliable native text layer. Details are provided in Appendix~\ref{app:samples-generation}.

\paragraph{Consistency verification.}

To filter hallucinations and schema violations, each generated sample is re-issued to a cross-family verifier pool (Qwen3-VL-235B-A22B-Thinking \cite{qwen3vl2025}, Qwen3.5-397B-A17B-FP8 \cite{qwen35blog} and Kimi K2.6 \cite{kimi2026k2_5}) sampling 3--5 responses at temperature 0.8. GPT-OSS 120B \citep{gptoss} judges them against the original under an equivalence relation that ignores cosmetic differences but flags factual disagreements. A sample is accepted when at least 2 verifiers agree in the majority of their responses, retaining 35--40\% of candidates; most rejections trace to residual PDF text-layer artifacts (Appendix~\ref{app:consistency-verification}).

\paragraph{Text refinement.} After initial post-training we observed degraded adherence to formatting instructions. The refinement stage rewrites annotations into target formats and augments prompts with explicit formatting rules via Qwen3.5-397B, covering common value types (numbers, dates, names) and applying auxiliary normalization (abbreviations, case) elsewhere, recasting the task from pure extraction into joint extraction and normalization (Appendix~\ref{app:text-refinement}).

\paragraph{Rendered document augmentation} 

External documents in our open-data corpus are pristine --- white pages, crisp glyphs, uniform illumination --- while production inputs are camera captures, low-DPI scans and screenshots. To close this gap we apply a three-stage pipeline at sample-serving time: photometric and geometric perturbations (contrast, brightness, sharpness, rotation, zoom, shear) modeling exposure and scanner deformations; elastic warps via a bilinear vector field, mimicking page curl; and spatially varying noise. The pipeline runs inside workers decoupled from training nodes (Appendix~\ref{app:dataserv}). Component ablation (Appendix~\ref{app:dadc-ablation}) confirms each stage is necessary: intermediate setups merely recover in-house increments, and only the full verified pipeline strictly improves model quality.

\section{Experiments}
\label{sec:experiments}

\paragraph{Experiment Setup}
We select Qwen3.5-35B-A3B-Base for its multilingual and visual strengths and favorable per-document cost (Section~\ref{sec:cost_analysis}), and fine-tune it for document understanding via SFT 
on 4 nodes with 8 H100 GPUs (hyperparameters in Appendix~\ref{app:hyperparams}). We benchmark against Qwen2.5-VL-72B-Instruct-AWQ~\citep{bai2025qwen25vl}, Qwen3.5-35B-A3B-Base, Qwen3.5-122B-A10B-FP8, Qwen3.5-397B-A17B-FP8~\citep{qwen35blog}, and Kimi K2.6~\citep{kimi2026k2_5} under identical inference conditions. Evaluation spans internal and public document benchmarks; online evaluation applies quality-adjusted cost analysis (Section~\ref{sec:cost_analysis}) to translate scores into per-document economics against the human-annotation baseline.

\paragraph{Deployment constraints} Production latency SLAs preclude test-time reasoning. Our model holds 95th-percentile latency well below 10-second our SLA across concurrency levels on a single H100, while both dense and MoE baselines breach it at 2--3 workers even on larger GPU allocations (Appendix~\ref{app:lat-analysis}). Cost analysis therefore uses non-reasoning inference, reasoning baselines in Table~\ref{tab:model_comparison} only bound headroom under relaxed latency.

\paragraph{Benchmarks}

Our primary offline evaluation is an internal suite of 12 benchmarks reflecting the production distribution, organized into three groups: \textit{Single-page (SP)} for one-page understanding over clean PDFs, in-the-wild scans and photos; \textit{Multi-page (MP)} for long-context comprehension over image sequences; and \textit{Fine-tuned (FT)} for business-critical workflows partially seen during training (details on data collection, curation, and contamination control are in Appendix~\ref{app:benchmarks}). Field extraction is evaluated by averaging exact accuracy~---all fields must match ground truth, directly proxying the full automation ratio—and field-level $F_1$, computed per field per sample, capturing incremental improvements invisible to exact accuracy; document classification uses standard accuracy and macro-$F_1$. 
We complement evaluation with the \textit{MWS Vision Benchmark}~\cite{mwsvisionbench2026} as a public Russian-language benchmark covering diverse business and technical documents, and report cross-lingual transfer results on widely used open benchmarks (Table~\ref{tab:external-bench}): OCRBench~\citep{liu2024ocrbench}, OCRBench~v2~\citep{fu2024ocrbenchv2}, CC-OCR~\citep{yang2024ccocr}, DocVQA~\citep{mathew2021docvqa}, and InfoVQA~\citep{mathew2022infographicvqa}.

\paragraph{Results}
Table~\ref{tab:model_comparison} reports results across all four benchmark groups. Our model achieves the best overall score, 0.767 averaged over non-fine-tuned benchmarks, outperforming the strongest open-source baseline, Qwen3.5-397B-A17B-FP8-Reasoning, by 2.4 absolute points despite being more than an order of magnitude smaller in active parameters, and improving over its base model Qwen3.5-35B-A3B-Base by 3.4 points, from 0.733 to 0.767, with consistent leads across single-page, fine-tuned, and open benchmarks and competitive multi-page performance at 0.764.

Two observations stand out. First, we attribute the strong multi-page score of Qwen3.5-397B-A17B-FP8-Reasoning, 0.777, to test-time scaling, where reasoning traces decompose dispersed visual evidence into intermediate steps; our model achieves comparable quality at 0.764 through SFT-acquired capabilities alone, without the latency and cost of reasoning-mode inference, as detailed in Section~\ref{sec:cost_analysis}. Second, excluding fine-tuned benchmarks to rule out in-distribution effects, our model still leads at 0.767 over 0.743, demonstrating generalization beyond the training distribution.

Production latency SLAs preclude reasoning-mode inference, so we treat such models as reference points rather than deployable candidates. Among deployable configurations, our model leads on every benchmark group, including multi-page at 0.764 over 0.745 for non-reasoning Qwen3.5-397B-A17B-FP8.

\begin{table*}
  \centering
  \resizebox{\textwidth}{!}{%
  \begin{tabular}{@{}lcccccc@{}}
    \toprule
    \textbf{Model} & \textbf{Single-page} & \textbf{MWS (val)} & \textbf{Multi-page} & \textbf{FT} & \textbf{Avg w/o FT} & \textbf{Avg} \\
    \midrule
    \texttt{Qwen2.5-VL-72B-Instruct-AWQ}      & 0.785 & 0.635 & 0.698 & 0.645 & 0.706 & 0.691 \\
    \texttt{Qwen3.5-35B-A3B-Base}             & \underline{0.827} & 0.677 & 0.697 & 0.722 & 0.733 & 0.730 \\
    \texttt{Kimi K2.6-Reasoning}              & 0.747 & 0.664 & 0.728 & \underline{0.886} & 0.713 & 0.756 \\
    \texttt{Qwen3.5-122B-A10B-FP8}            & 0.794 & 0.637 & 0.713 & 0.745 & 0.715 & 0.722 \\
    \texttt{Qwen3.5-397B-A17B-FP8}            & 0.726 & 0.674 & 0.745 & 0.770 & 0.715 & 0.729 \\
    \texttt{Qwen3.5-397B-A17B-FP8-Reasoning}  & 0.757 & \underline{0.695} & \textbf{0.777} & 0.815 & \underline{0.743} & \underline{0.761} \\
    \textbf{Our model}                        & \textbf{0.842} & \textbf{0.700} & \underline{0.764} & \textbf{0.956} & \textbf{0.767} & \textbf{0.814} \\
    \bottomrule
  \end{tabular}%
  }
  \caption{
    Comparison of models across evaluation settings. Reasoning-mode inference; reported for reference only.
  }
  \label{tab:model_comparison}
\end{table*}

\paragraph{How much of SFT gain was already in base model?}
\label{sec:sft_unlock} 

SFT may either refine behavior already present in the base model or unlock latent capacity that pre-training instilled but left dormant in the zero-shot setup. The distinction is operational: refinement saturates quickly with data, whereas unlocking keeps targeted post-training high-leverage at relatively modest volumes.

We compare two base models, Qwen3.5-9B-Base and Qwen3.5-35B-A3B-Base, post-trained on an identical mixture under their respective best hyperparameter setups (Table~\ref{tab:sft_unlock}).

\begin{table}[!htbp]
  \centering
  \small
  \resizebox{\columnwidth}{!}{%
  \begin{tabular}{@{}lccccc@{}}
    \toprule
    \textbf{Model} & \textbf{SP} & \textbf{MWS} & \textbf{MP} & \textbf{FT} & \textbf{Avg w/o FT} \\
    \midrule
    9B-Base          & 0.780 & 0.588 & 0.626 & 0.640 & 0.667 \\
    9B-Base + SFT      & \textbf{0.808} & \textbf{0.620} & \textbf{0.718} & \textbf{0.955} & \textbf{0.714} \\
    \midrule
    35B-A3B-Base     & 0.827 & 0.677 & 0.697 & 0.722 & 0.732 \\
    35B-A3B-Base + SFT      & \textbf{0.842} & \textbf{0.700} & \textbf{0.764} & \textbf{0.956} & \textbf{0.767} \\
    \bottomrule
  \end{tabular}%
  }
  \caption{SFT effect across base scales. SP: single-page, MP: multi-page, FT: fine-tuned.}
  \label{tab:sft_unlock}
\end{table}

The gains are disproportionately large relative to the SFT corpus, which is small by pre-training standards, and they generalize beyond SFT coverage, with consistent improvements across all benchmark groups. Despite identical SFT data, the models do not meet at a common ceiling: post-trained Qwen3.5-35B-A3B-Base retains its lead across all groups, suggesting pre-training capacity sets the upper bound while SFT controls how much is realized.

\paragraph{Is internal data a scaling factor?}
\label{sec:internal_scaling}

Internal and open-domain data play different roles: internal data is aligned with the production distribution but bounded by templates and a fixed field taxonomy, while open-domain data offer unbounded diversity at the cost of aggressive curation (Section~\ref{sec:dadc}). We assess their relative scaling via controlled ablations over the mixing ratio at fixed training budget (Table~\ref{tab:internal_scaling}).
\begin{table}[!htbp]
  \centering
  \small
  \resizebox{\columnwidth}{!}{%
  \begin{tabular}{@{}lccccc@{}}
    \toprule
    \textbf{Mixture} & \textbf{SP} & \textbf{MWS} & \textbf{MP} & \textbf{FT} & \textbf{Avg w/o FT} \\
    \midrule
    Base (no SFT)    & 0.827 & 0.672 & 0.697 & 0.722 & 0.729 \\
    Internal only    & \underline{0.840}   & 0.685   & 0.738   & 0.951   & 0.754   \\
    Open only        & 0.835   & \textbf{0.700}   & 0.677   & 0.793   & 0.737   \\
    1:1 mix          & 0.837   & 0.689   & \textbf{0.768}   & \underline{0.955}   & \underline{0.764}   \\
    1:4 (open-heavy) & \textbf{0.842} & \textbf{0.700} & \underline{0.764} & \textbf{0.956} & \textbf{0.767} \\
    \bottomrule
  \end{tabular}%
  }
  \caption{Effect of internal:open data ratio on downstream quality (fixed training budget).}
  \label{tab:internal_scaling}
\end{table}
Internal data gains saturate quickly, reflecting low template diversity in production workflows. Open-domain data scales further, with performance improving well beyond the internal-data plateau, driven by CC diversity and DADC filtering. A hybrid mixture dominates either source alone, with the open-heavy ratio yielding the best trade-off. Beyond scaling, internal data offer the highest-fidelity signal for early-phase discovery and failure-mode diagnosis on production-representative inputs. We accordingly adopt a hybrid strategy: internal data drives early exploration and targeted gap-filling, while open-domain synthetic data carries the bulk of the scaling effort.

\paragraph{Cross-lingual transfer on public benchmarks}
Our model improves over the base on every benchmark, with the largest gains on DocVQA, from 82.14 to 90.10, and InfoVQA, from 51.10 to 66.30, confirming that DADC transfers to English despite a predominantly Russian training corpus. Smaller gains on OCRBench, +15\%, and OCRBench v2, +1.1\%, are notable given their diversity: recognition, referring, spotting, relation extraction, parsing, math, and reasoning, of which structured extraction is a small fraction, indicating generalization beyond DADC's direct training signal. CC-OCR shows a slight aggregate regression, from 81.74 to 80.40, but on slices closest to our deployment we improve: KIE rises from 90.39 to 92.81 and Document Parsing from 61.92 to 63.77. A meaningful gap to Qwen3.5-397B-A17B-FP8 persists, most visibly on InfoVQA, 66.30 against 89.58, reflecting the cost of an order-of-magnitude reduction in active parameters.

\begin{table}[h]
  \centering
  \small
  \resizebox{\columnwidth}{!}{%
  \begin{tabular}{lccc}
    \toprule
     & \textbf{Base} & \textbf{Ours} & \textbf{Qwen3.5 397B} \\
    \midrule
    OCRBench        & 848   & \underline{863}   & \textbf{889} \\
    OCRBench v2 (en)& 55.1  & \underline{56.2}  & \textbf{61.1} \\
    CC-OCR          & \underline{81.74} & 80.40 & \textbf{84.86} \\
    DocVQA (test)   & 82.14 & \underline{90.10} & \textbf{96.74} \\
    InfoVQA (test)  & 51.10 & \underline{66.30} & \textbf{89.58} \\
    \bottomrule
  \end{tabular}%
  }
  \caption{Performance on English-language public document understanding benchmarks (non-reasoning mode).}
  \label{tab:external-bench}
\end{table}

\section{Quality-adjusted Cost Analysis}
\label{sec:cost_analysis}

A higher-scoring model may cost more to serve than the annotation it displaces, so our quality-adjusted cost framework translates per-field quality into per-document economics, enabling unified comparison across models and the manual baseline.

\paragraph{Inference Cost}

Consider a model served on $G_m$ H100 GPUs at utilization $u \approx 0.5$, calibrated to peak-hour SLAs, with monthly GPU cost $C_{\mathrm{GPU}}$ and $T_{\mathrm{month}} = 30 \cdot 24 \cdot 3600$ seconds, the per-document inference cost is

\begin{equation}
  y(m,w) = \frac{G_m \cdot C_{\mathrm{GPU}}}
                 {u \cdot T_{\mathrm{month}} \cdot \mathrm{RPS}_m(w)},
  \label{eq:inference_cost}
\end{equation}

\noindent where $\mathrm{RPS}_m(w)$ is the profiled throughput for model--workflow pair. Because $G_m$ ranges from $1$ for compact MoEs to $8$ for dense 400B models, it is the primary driver of cost differences.

\paragraph{Field-level Routing} A single aggregate score obscures the per-field variation that governs deployment economics.  We develop the framework for field extraction, the dominant cost driver, but it extends to classification and criteria validation by treating each decision as a field. Given per-field accuracy $a_f$ and precision $p_f$ on a held-out subset, each field is routed into one of three modes via thresholds $\theta_{\mathrm{auto}}$ and $\theta_{\mathrm{assist}}$: fully automated ($a_f \geq \theta_{\mathrm{auto}}$), where the prediction is accepted at inference cost only; human-assisted ($a_f < \theta_{\mathrm{auto}} \land p_f \geq \theta_{\mathrm{assist}}$), where the prediction is surfaced as a pre-fill for the operator to confirm or correct; and fully manual ($a_f < \theta_{\mathrm{auto}} \land p_f < \theta_{\mathrm{assist}}$), where the prediction is discarded and the field is annotated from scratch. Routing is fixed per field and per workflow at deployment time and remains unchanged until the next model update, different fields within the same document may fall into different modes.

\paragraph{Cost Model}

The model rests on two atomic units: $x$,  the fully loaded cost of manual field annotation, covering operator time, supervision, quality control, and $y(m,w)$, the per-document inference cost from Eq.~\ref{eq:inference_cost}. Assisted-mode outcomes carry asymmetric costs: confirmation costs $0.3x$, reflecting the reduction from search-and-transcribe to read-and-confirm, while correction costs $1.1x$, the overhead arising from error detection and the context switch of discarding a displayed suggestion (coefficient calibration in Appendix~\ref{app:cost-coeff-calibration}). Since the confirmation-correction split for an assisted field $f$ is governed by its precision $p_f$, the expected extraction cost is
\begin{equation}
  c_{\mathrm{a}}(f) = 0.3x \cdot p_f + 1.1x \cdot (1 - p_f).
  \label{eq:assist_cost}
\end{equation}
Eq.~\ref{eq:assist_cost} decreases monotonically in $p_f$, and the total per-document cost for workflow $w$ under model $m$ is
\begin{equation}
  C_w(m) = y(m,w) + x \cdot \vert \mathcal{F}^{\mathrm{h}}_w \vert + \sum_{f \in \mathcal{F}^{\mathrm{a}}_w} c_{\mathrm{a}}(f),
\label{eq:total_cost}
\end{equation}
\noindent where $\mathcal{F}^{\mathrm{a}}_w$ and $\mathcal{F}^{\mathrm{h}}_w$ are the sets of fields per assisted and manual modes. Against the fully manual baseline $C_w(\mathrm{baseline}) = x \cdot \vert \mathcal{F}_w \vert$, the cost reduction ratio is
\begin{equation}
  \rho_w(m) = 1 - \frac{C_w(m)}{C_w(\mathrm{baseline})}.
  \label{eq:cost_reduction}
\end{equation}
The deployment is economically justified when $\rho_w(m) > 0$.
The automation threshold $\theta_{\mathrm{auto}} = 0.95$ is a business decision bounding the residual error rate on fully automated fields. The assistance threshold $\theta_{\mathrm{assist}}$ is derived analytically from the break-even condition $c_{\mathrm{assist}}(f) = x$, yielding $\theta_{\mathrm{assist}} = 0.125$: below this precision, correcting a likely wrong suggestion costs more than annotating from scratch, so the prediction is discarded.
\paragraph{Results} Table~\ref{tab:cost_comparison} reports per-document inference cost and cost reduction ratio across two workflow groups differing in field count ($w_1$: 12 fields on average, $w_2$: 6) and per-field accuracy; all configurations are benchmarked on 8 H100s, with $G_m$ denoting the minimum footprint per replica. Despite competitive offline quality, large-model serving costs push $\rho_w$ below zero: compute exceeds the displaced annotation effort. Qwen3.5-397B illustrates this clearly on both groups. As field count decreases, the fixed serving cost becomes harder to amortize, and large models' $\rho_w$ deteriorates sharply; smaller models degrade gracefully. Our model sustains nearly 5$\times$ higher throughput than Qwen2.5-VL-72B and yields the highest $\rho_w$ in both groups, closing the cost--quality gap that motivates this work.

\begin{table}[t]
  \centering
  \small
  \resizebox{\columnwidth}{!}{%
  \begin{tabular}{@{}lcccc@{}}
    \toprule
    \textbf{Model} & $G_m$ & Throughput & $\rho_{w_1}(m)$ & $\rho_{w_2}(m)$\\
    \midrule
    \texttt{Qwen3.5-397B-A17B-FP8}       & 8   & $0.025\times$ & -0.115 & -1.582  \\
    \texttt{Qwen2.5-VL-72B-Instruct-AWQ} & 1   & $0.200\times$ & 0.563  & 0.534 \\
    \texttt{Qwen3.5-35B-A3b-Base}        & 1   & $1.000\times$ & 0.691 & 0.791 \\
    \textbf{Our model}                   & 1   & $1.000\times$ & \textbf{0.819}  & \textbf{0.861} \\
    \bottomrule
  \end{tabular}%
  }
  \caption{
    Throughput (normalized to our model) and cost reduction ratio across two production workflow groups measured at an average workload of 6k input tokens (including 5k visual) and up to 0.5k output tokens.
  }
  \label{tab:cost_comparison}
\end{table}

\section{Conclusion}

Deploying a VLM at scale in a regulated setting is gated not by benchmark scores but by per-document economics. We presented a 35B-A3B VLM, domain-adapted via a difficulty-aware curation pipeline on open-domain PDFs mixed with in-house production data, that outperforms baselines an order of magnitude larger on our production benchmarks. Benchmark-leading quality is necessary, but not sufficient. Our quality-adjusted cost framework makes this trade-off explicit, identifying our model as the most economically viable solution. We believe the paradigm offers a broadly useful template for grounded model selection in deployment economics.

\section*{Limitations}

\paragraph{Reasoning and reinforcement learning.} Our corpus centers on direct field extraction and lacks multi-step reasoning examples, such as reconciling information across many pages or resolving cross-referential dependencies. The model may therefore underperform on workflows requiring compositional reasoning or long-range evidence aggregation. We deliberately chose not to pursue an reinforcement learning (RL) stage at this point: our SFT pipeline already matches or exceeds reasoning-mode variants of substantially larger models on the benchmarks evaluated in Table~\ref{tab:model_comparison}, suggesting that the current synthetic-data recipe captures a large share of the extractable signal on our present evaluation suite without an explicit RL objective. Moreover, RL with verifiable rewards requires that each training example admit a ground-truth answer. In our setting this is a harder bar than SFT supervision, which tolerates noisier labels: either domain experts must curate multi-step problems with unambiguous solutions, or the synthetic-data pipeline must be redesigned to produce tasks whose answers are verifiable by construction. We regard this as a natural next step once returns from SFT data scaling diminish.

\paragraph{Cross-lingual coverage.} Although training data includes multiple Cyrillic languages alongside Russian, we have not run a per-language ablation, leaving cross-lingual transfer behavior unclear.

\paragraph{Post-training ceiling.} SFT has not pushed the model to its performance ceiling, a gap we attribute primarily to pre-training limitations rather than the post-training recipe. We plan to continue scaling post-training --- expanding data diversity, reasoning coverage, and supervision quality --- to push performance further.

\section*{Ethical Statement}

\paragraph{Human annotation.} All annotation, quality control, and benchmark construction were performed by trained in-house operators employed under standard full-time contracts, with compensation aligned with local labor regulations. No external crowdsourcing was used. The retrospective study calibrating the assisted-mode cost coefficients (Appendix~\ref{app:cost-coeff-calibration}) used aggregated, anonymized operational telemetry from 50 annotators over three months; no individual-level evaluation or disciplinary action was derived from it. The framework is explicitly designed around human-in-the-loop operation: lower-confidence fields are routed to assisted or manual modes, preserving operator oversight, and deployment is coordinated with workforce planning to redirect capacity toward higher-complexity tasks.

\paragraph{Data Privacy.} Documents processed by the platform contain PII. This constraint directly motivated on-premise fine-tuning and deployment rather than reliance on external APIs. All internal data was stored within the institution's secured infrastructure under role-based access control, encryption at rest and in transit, and auditable access logs, as mandated for regulated PII data. No internal documents, derived training samples, or checkpoints trained on internal data were transferred outside the secured perimeter or used to query third-party services. Cross-model verification of synthetic samples used open-weight models executed within the same perimeter and was applied exclusively to the open-domain corpus.

\paragraph{Responsible Use.} The model is intended for document understanding within a regulated enterprise setting and is not suitable for unconstrained consumer-facing deployment without further safety review. The cost framework is calibrated to a specific operational setting and should be re-calibrated before transfer to other environments. Routing decisions must be revisited whenever the document distribution, schema, or workflow changes materially. 
\bibliography{custom}

\newpage

\appendix

\section{Benchmark Suite Details}
\label{app:benchmarks}

\subsection{Dataset Composition}

Table~\ref{tab:benchmark_details} enumerates the internal benchmarks grouped by task type. Each group spans the three canonical task families encountered in production: field extraction, document classification, and criteria validation. Each benchmark item was independently annotated by 3 of 5 trained annotators. Disagreements were resolved through adjudication, and items without majority consensus were excluded.

\begin{table*}
  \centering

  \resizebox{\textwidth}{!}{%
  \begin{tabular}{@{}lllr@{}}
    \toprule
    \textbf{Group} & \textbf{Subbenchmark} & \textbf{Task Type} & \textbf{Samples} \\
    \midrule
    \multirow{7}{*}{\textbf{Single-page (SP)}}
      & Payment Screenshot Amounts       & Field Extraction       & 2{,}000 \\
      & B2B Invoices                     & Criteria Validation, Field Extraction & 2{,}000 \\
      & Document Type Classification     & Document Classification & 1{,}909 \\
      & Short Court Rulings              & Field Extraction       & 1{,}524 \\
      & ID Documents                     & Field Extraction       & 900 \\
      & Income Certificates              & Field Extraction       & 610 \\
      & Restaurant Receipts              & Field Extraction       & 410 \\
    \midrule
    \multirow{3}{*}{\textbf{Multi-page (MP)}}
      & Enforcement Documents            & Field Extraction       & 3{,}376 \\
      & Banking Legal Documents          & Field Extraction       & 1{,}900 \\
      & Company Compliance Fillings      & Field Extraction       & 1{,}000 \\
    \midrule
    \multirow{2}{*}{\textbf{Fine-tuned (FT)}}
      & Retail Invoices                  & Field Extraction       & 1{,}911 \\
      & Court Case Files                 & Field Extraction       & 1{,}341 \\
    \bottomrule
  \end{tabular}
  }
  \caption{Internal benchmark suite, organized by evaluation group.}
  \label{tab:benchmark_details}
\end{table*}

\subsection{Contamination Control}
\label{sec:contamination}
For fine-tuned benchmarks, we partition documents by unique identifying entities --- company names, personal names, taxpayer identification numbers, and account numbers --- ensuring no entity appearing in the evaluation set is present in the training corpus. This guarantees that the reported metrics reflect generalization to unseen parties rather than memorization of specific document instances. Template-level contamination is likewise precluded: the fine-tuned benchmarks are drawn from workflows that are different from those represented in the training corpus, with disjoint document classes, layouts, issuing authorities, and field taxonomies. At all stages we perform MinHash deduplication against benchmarks to prevent data leakage, the prompts for evaluation data were written and curated by professional annotators and only lightly adapted per baseline to enforce a JSON-structured response; no per-baseline capability tuning was performed.

As a result, neither the surface form nor the structural skeleton of evaluation documents is observed during training, and the reported metrics reflect generalization across both entities and templates.

\paragraph{Distinguishing Court Case Files from Short Court Rulings.}

Although both originate from the judicial domain, the two benchmarks correspond to structurally distinct document classes. Court Case Files are multi-page interlocutory proceedings with schemas centered on procedural metadata, whereas Short Court Rulings are single-page terminal decisions from small-claims cases, with schemas centered on final disposition (outcome, awarded amount, ruling date). The two differ in document length, procedural stage, field taxonomy, and source workflow. Entity-level partitioning further guarantees no shared parties, case numbers, or taxpayer identifiers between the two, ensuring that performance in Short Court Rulings reflects genuine generalization to an out-of-distribution judicial document type.

\section{Internal Data Construction}
\label{app:internal}

\subsection{Prompt Template}
\label{app:internal-prompt}

Each training instruction is composed of a task description and an output schema. Figure ~\ref{fig:prompt-extraction} is the generalized template used for field-extraction tasks.

To mitigate template overfitting, each prompt component is drawn from a pool of semantically equivalent phrasings produced manually and augmented with Claude Sonnet 4.5 \citep{sonnet4_5}. The order of key-value pairs in the schema is randomly permuted across samples.

\begin{figure*}[!htbp]
\centering
\begin{minipage}{0.98\linewidth}
\begin{Verbatim}[fontsize=\small, frame=single, breaklines=true, breakanywhere=true]
You are an expert in {domain} documents and {task_type}. 
Your task is to extract precise values for a strictly defined list of entities from the document image. 
All specified entities must be included in the response, even if they are absent from the text.

Instructions:
1. Review the entity descriptions provided after "|" following entity names -- they clarify what to extract. 
   DO NOT include them in the response.
2. Extract the value for each entity from the document text.
3. Verify that the list of entities in your response matches the list below exactly -- no additions or 
   omissions.
4. Format your response strictly as JSON, where keys are field names and values are the extracted fields. 
   If a value for any entity is absent, set it to "nan".

Entity list:
FIELD_1 | {description}. Location: {location_cue}.
          Identifying features: {features}.
          Example: "{example_context}" -> extracted "{example_value}".
...
FIELD_N | {description}. Format: {format_constraint}.
Processing rules:
1. NOT ALL FIELDS MAY BE PRESENT IN THE DOCUMENT.
2. DO NOT HALLUCINATE OR SUBSTITUTE INCORRECT DATA.
3. FOR FIELDS ABSENT FROM THE TEXT, ALWAYS SET THE VALUE TO nan.
4. RETURN ONLY THE JSON RESPONSE, NO ADDITIONAL TEXT.

Return response in the following format:
{"FIELD_1": "...", "FIELD_2": "...", ..., "FIELD_N": "..."}
\end{Verbatim}
\end{minipage}
\caption{Prompt template for field
extraction from internal production data.}
\label{fig:prompt-extraction}
\end{figure*}

\section{Open Data Construction}
\label{app:synthetic}

\subsection{Data}
\label{app:open-data-eda}

The open-domain corpus is subsampled from a large-scale Common Crawl PDF dataset, retaining only documents identified as Russian, Belarusian, Ukrainian, or Kazakh via a GlotLID~\citep{kargaran2023glotlid} language identification model. The resulting collection comprises approximately 300K PDF documents (5M pages), predominantly Russian-language, selected to align with the operational deployment setting of the model.

\subsection{Text Layer Validation}
\label{app:text-layer}

PDFs are a compelling source of training data, but text extraction is fundamentally fragile: the PDF standard encodes glyph-positioning instructions whose mapping to characters depends on per-font tables (Encoding, ToUnicode CMap) that may be absent, incorrect, or deliberately obfuscated. In addition, a substantial portion of real-world documents embed text as raster images overlaid with a missing or corrupted text layer, causing structure-based extraction to return empty or garbled output. 

We address this with a \emph{dual-channel annotation scheme}. For each page, we (i) parse the PDF content stream to extract the embedded text layer and (ii) run an internal OCR engine on the rasterized page image. The embedded text layer, when valid, faithfully reproduces characters but is vulnerable to PDF-level artifacts (malformed encodings, hidden text objects). OCR, conversely, is robust to such structural anomalies, but introduces its own character- and symbol-level recognition errors. For each page we compute a consistency score between the two channels, using normalized character error rate with domain-specific tokenization. Pages whose alignment exceeds a fixed threshold are accepted; documents falling below it are discarded entirely --- such mismatches typically indicate either malformed font encodings or a scanned page whose text layer was itself generated by low-quality OCR at digitization time. 

A particularly subtle failure mode arises from documents containing a \emph{transparent OCR text layer} --- a byproduct of digitization pipelines (e.g., ABBYY FineReader) that overlay their own OCR output as invisible glyphs on top of the scanned image. Structurally indistinguishable from a high-confidence native text layer, such pages would otherwise poison training with compounded OCR artifacts. We proactively detect and exclude these pages, ensuring the final corpus contains only natively generated text layers verified against an independent OCR signal. A conceptually related strategy is employed by \citet{poznanski2025olmocr}, that uses native PDF text as an anchor signal to verify VLM output.

\paragraph{Verification statistics.} Applied to the open-data corpus (Appendix~\ref{app:open-data-eda}), the procedure retains approximately 40\% of input pages, with matched precision $= 0.80$, matched recall $= 0.90$, and average normalized Levenshtein distance $= 0.05$ over matched spans. The remaining 60\% are discarded as malformed text layers or scanned pages whose embedded text was itself produced by low-quality OCR at digitization time. In future work, we plan to recover signal from currently rejected pages, either by re-OCRing them with a strong engine, or by using them directly with a VLM to produce the resulting annotations.

\subsection{Document Sampling}
\label{app:document-sampling}

Pages that pass text-layer validation are dominated by continuous prose --- a poor match for our target deployment, where models must extract fields from forms, tables, and dense structured layouts. To rebalance the corpus toward visually challenging documents, we apply a two-stage filter: a fine-grained taxonomic classifier whose non-\textsc{PLAIN\_TEXT} labels directly admit a page, followed by a fact-extractability scorer that rescues structurally sparse pages that nonetheless contain extractable facts.

\paragraph{Stage 1: Taxonomic classification.} Each candidate page is scored by Qwen2.5-VL-72B-Instruct-AWQ against a fixed taxonomy of visual-reasoning content. The full prompt is shown in Figure~\ref{fig:taxonomy-prompt}. Pages assigned at least one non-\textsc{Plain\_Text} label are admitted directly to the candidate pool.

Across successfully classified pages, Table~\ref{tab:taxonomy-distr} reports the distribution over primary categories, and Table~\ref{tab:stage2_scores} the distribution of complexity scores.

\begin{figure*}[!htbp]
\centering
\begin{minipage}{0.98\linewidth}
\begin{Verbatim}[fontsize=\small, frame=single, breaklines=true, breakanywhere=true]
Analyze this document page image and classify it according to the visual-reasoning content it contains. 
Your goal is to identify pages whose layout supports non-trivial information extraction or inference.

**TAXONOMY** Assign up to three labels from the following set:

- CHARTS: charts, plots, bar/line/pie graphs, numeric figures whose interpretation requires reading values off axes 
  or legends.
- LOGIC_DIAGRAMS: flowcharts, decision trees, state machines, BPMN diagrams, any directed-graph structure encoding
  control flow or rules.
- SCHEMATIC: floor plans, circuit diagrams, engineering layouts, any content whose meaning depends on 2D spatial 
  arrangement.
- INFOGRAPHIC: mixed-modality compositions combining icons, short text blocks, and stylized visuals to convey 
  information.
- TABULAR: structured data in rows and columns, including:
   * Traditional tables with visible borders/gridlines
   * Borderless tables (data aligned by spacing/tabs)
   * Matrices, grids, comparison charts, financial statements,
     price lists, schedules in table format
   * Spreadsheet-like data presentations
  EXCLUDE from TABULAR: single-column bulleted/numbered lists, paragraphs of regular text, isolated key-value pairs, 
  forms with scattered input fields.
- PLAIN_TEXT: continuous prose, standard book/article layout, decorative pages, or any page that contains none of 
  the above. PLAIN_TEXT is MUTUALLY EXCLUSIVE with all other categories.
  
Output schema:
{
  "categories": [...],  // up to 3 labels
}
\end{Verbatim}
\end{minipage}
\caption{Stage~1 taxonomic-classification
prompt. The classifier assigns up to three labels per page from a fixed
taxonomy; \textsc{Plain\_Text} routes the page to Stage~2.}
\label{fig:taxonomy-prompt}
\end{figure*}

\begin{table}[!htbp]
  \centering
  \small
  \begin{tabular}{lr}
    \toprule
    \textbf{Category} & \textbf{Share} \\
    \midrule
    \textsc{Plain\_Text}      & 72.9\% \\
    \textsc{Tabular}          & 23.3\% \\
    \textsc{Schematic}        & 2.3\%  \\
    \textsc{Charts}           & 0.6\%  \\
    \textsc{Logic\_Diagrams}  & 0.4\%  \\
    \textsc{Infographic}      & 0.4\%  \\
    \bottomrule
  \end{tabular}
  \caption{Distribution of taxonomic labels,
  counted among all assigned labels across pages.}
  \label{tab:taxonomy-distr}
\end{table}

\begin{table}[!htbp]
  \centering
  \small
  \begin{tabular}{lrr}
    \toprule
    \textbf{Score} & \textbf{Visual} & \textbf{QA Potential} \\
    \midrule
    0  & 0.00\%  & 0.00\%  \\
    1  & 0.61\%  & 0.48\%  \\
    2  & 51.68\% & 54.12\% \\
    3  & 10.20\% & 18.12\% \\
    4  & 16.57\% & 10.94\% \\
    5  & 9.80\%  & 6.59\%  \\
    6  & 8.33\%  & 4.38\%  \\
    7  & 1.93\%  & 3.36\%  \\
    8  & 0.51\%  & 1.61\%  \\
    9  & 0.37\%  & 0.40\%  \\
    10  & 0.00\%  & 0.00\%  \\
    \bottomrule
  \end{tabular}
  \caption{Distribution of Stage~2 visual structure and QA potential scores across pages routed to the fact-extractability filter.}
  \label{tab:stage2_scores}
\end{table}

\paragraph{Stage 2: Fact-extractability rescue.} \textsc{PLAIN\_TEXT} pages are passed through a second filter that scores each page on two axes --- visual structure complexity and QA potential --- framed around the question: \emph{can a precise, unambiguous question be asked about this document?} Pages whose QA potential score $\geq 5$ \emph{or} visual score $\geq 5$ are admitted to the candidate pool; the rest are discarded. The full prompt is shown in Figure~\ref{fig:fact-extract-prompt}.

The motivation for this stage is twofold. First, the Stage~1 taxonomy is deliberately narrow: it targets pages whose \emph{visual structure} encodes the answer (tables, diagrams, charts). Yet many \textsc{Plain\_Text}-labeled pages --- receipts with stamps, simple invoices, short forms with handwritten fields, dense administrative documents --- contain concrete extractable facts (dates, amounts, identifiers, named entities) without exhibiting any of the Stage~1 visual structures. Second, Stage~1 alone discards roughly 70\% of all classified pages as \textsc{Plain\_Text} (Table~\ref{tab:taxonomy-distr}), making the funnel uncomfortably sharp; routing these pages through a second, content-oriented filter recovers meaningful signal that would otherwise be lost wholesale. Stage~2 thus complements Stage~1 rather than relaxing it: scoring is conditioned on the assumption that the page already failed the structural filter, and admission requires that at least one of the two axes register an unambiguous positive signal.

\begin{figure*}[!htbp]
\centering
\begin{minipage}{0.98\linewidth}
\begin{Verbatim}[fontsize=\small, frame=single, breaklines=true, breakanywhere=true]
You are a strict Data Curator selecting documents for VLM training on Information Extraction. Your filter: "Can a precise question with an unambiguous answer be asked about this document?"

KEEP:
- Documents where the answer is a concrete substring in the image (date, amount, name, table cell value, code).
- Documents with complex but structured layouts.
DISCARD:
- Continuous prose (articles, letters, contracts without tables) that admits only general questions ("What is this about?", "Summarize").
- Documents where questions would be artificial.
SCORING:
A. Visual Structure (0-10):
   - 0-3: continuous text, standard book layout.
   - 4-7: lists, bold headers, simple key-value forms.
   - 8-10: complex tables, forms, receipts, stamps, handwritten fields.
B. QA Potential (0-10):
   - 0-3: descriptive/literary text. Only "what is this about?" possible.
   - 4-7: 2-3 concrete facts (e.g., date + document number).
   - 8-10: dense with extractable facts (5+ unambiguous questions).
   
PROCEDURE:
1. Mentally formulate 3 strict questions.
2. If only "what is this about?" works, score low.
3. Score by data density.
Return JSON with: reasoning, data_types, fact_check, visual_score,
qa_potential_score, decision (KEEP if qa_potential_score >= 5 OR
visual_score >= 5, else DISCARD).
\end{Verbatim}
\end{minipage}
\caption{Stage~2 fact-extractability prompt. Applied only to \textsc{Plain\_Text} pages from Stage~1.}
\label{fig:fact-extract-prompt}
\end{figure*}

The union of these two stages forms the final candidate pool from which the synthetic corpus is sampled.

\subsection{Samples Generation}
\label{app:samples-generation}

For each page admitted to the candidate pool, we synthesize a set of samples. By default, generation uses a single prompt that conditions on the page's extracted text (with page-boundary markers preserved) and produces multi-field information extraction questions. For pages whose Stage~1 taxonomy includes a \textsc{Tabular} label, we additionally apply a table-oriented prompt that targets row/column intersections, conditional lookups, and header--cell relations. The general prompt is shown in Figure~\ref{fig:general-gen-prompt}. Generation is performed by both Qwen3.5 397B and GLM5.1 \citep{glm5_2026} conditioned on the page text. Across both prompts, we enforce strict OCR-mode extraction: values are returned verbatim from the source text without normalization, case folding, or punctuation changes, and all numeric fields (sums, codes, identifiers, requisites) are returned as strings to preserve leading zeros and original formatting that will be refined in next stages.

\begin{figure*}[!htbp]
\small
\begin{minipage}{0.98\linewidth}
\begin{Verbatim}[fontsize=\small, frame=single, breaklines=true, breakanywhere=true]
Your task is to generate a set of Information Extraction questions for the document, focused on 
**document understanding** and **targeted extraction of data spanning multiple pages** under given conditions. 
Answers must be taken strictly from the text. Return the result as a JSON array.

Input:
*   document_text: the full document text (including page-boundary
    markers).
*   n_questions: number of questions to generate.

Questions must test the model's ability to find answers located across multiple pages and to extract key 
information from a document. Questions should be complex and diverse, covering as many 
different data-retrieval scenarios as possible.

Question types (in priority order):
1.  **Multi-page entity or process extraction** -- e.g., extracting a sequence of events in a
    court case, or information about the court that is spread across several pages.
2.  **Requisite extraction** -- tasks requiring point-wise retrieval of specific values from the
    text (dates, surnames, amounts, etc.).
3.  **Document-understanding tasks** -- simple yes/no questions, or questions probing what the
    document is about.
Format requirements:
*   **"question"**: the task statement + the expected JSON schema (keys and types). The schema must 
    be flat (when extracting a single object); do not use questions that require nested answers.
*   **"answer"**: a string containing valid JSON. Values must be taken strictly from document_text.

Suggested generation procedure:
  1. Describe what the document is about and what data it contains.
  2. List the entities present (people, organizations, etc.).
  3. Describe the sequence of events in the document and how the listed entities participate in 
     them.
  4. Based on the event sequence and entity descriptions, formulate the extraction task or pose
     questions. Tasks should be varied; you may add descriptive context, but must not reveal
     the answer (bad example: "Find the client's registration address, which starts with
     Pushkin St., bldg. 7").
  5. Respond in valid JSON inside a ```json ... ``` block, so the answer can be parsed
     via json.loads().
     
Additional requirements:
  1. Extract data in OCR mode: no normalization, no corrections, no changes to case, punctuation,
     or whitespace -- strictly as it appears in document_text.
  2. All numeric values (sums, codes, numbers, articles, requisites, etc.) MUST be returned as 
     STRINGS, preserving leading zeros and original formatting.
  3. EACH "question" MUST REQUIRE EXTRACTING NO LESS THAN FOUR AND NO MORE THAN EIGHT FIELDS.
     THIS IS A STRICT RULE; THE NUMBER OF FIELDS MUST VARY ACROSS QUESTIONS.
  4. The JSON-answer format inside each question must parse with json.loads() without errors.
  5. The task should be well-described, not just ("Extract these entities", "Find these entities").
     Phrasing must be varied and imitate realistic human queries.
  6. Do not use "" in answers. If information is missing, write "N/A".

Response format:
```json
[
  {
    "question": "Question text. JSON answer format: { ...schema... }",
    "answer": "{ ...json answer... }"
  }
]
  n_questions: 4
  document_text:
\end{Verbatim}
\end{minipage}
\caption{\label{fig:general-gen-prompt} General instruction-generation prompt (translated to English), applied to every admitted page. Targets multi-page, multi-field information extraction questions in OCR-faithful mode.}
\end{figure*}

\paragraph{Multi-page document understanding} Production workflows routinely require reasoning over documents that span up to 20 pages, and our open-data generation explicitly targets both regimes in which such context is consumed. The first is \emph{needle-in-a-haystack} retrieval, where a single field or fact must be located within a predominantly irrelevant context; the second is \emph{multi-page analysis}, where the answer is synthesized from evidence scattered across several pages. Question generation is conditioned on page-boundary markers preserved in the extracted text, allowing the generator to construct queries whose answers deliberately cross page breaks. This dual focus ensures the model is trained to both isolate local evidence under heavy distractor load and to aggregate non local evidence across the document.

\subsection{Consistency Verification}
\label{app:consistency-verification}

Samples produced by the generation stage (Appendix~\ref{app:samples-generation}) are not used directly: single-shot outputs from a frontier models still exhibit a long tail of hallucinated fields, misread digits, and silent schema violations. To filter these out, we apply a consistency-verification stage based on cross-model agreement.

\paragraph{Verifier pool.} For each generated pair we re-issue the question against a pool of three independent VLMs --- Qwen3-VL-235B-A22B-Thinking, Qwen3.5-397B-A17B-FP8, and Kimi K2.6 --- each sampling 3--5 responses at temperature 0.8. The pool is chosen for diversity across model families and pre-training data; agreement among its members is a stronger signal than agreement within a single model's sampling distribution.

\paragraph{LLM-as-Judge.} The original generated answer and the verifier-pool responses are passed to a judge model GPT-OSS 120B \citep{gptoss}. The judge compares responses field-by-field under a fixed equivalence relation that ignores cosmetic differences (formatting, casing, line-break artifacts, JSON key order, date-format variants) but flags factual disagreements, hallucinated values, wrong-field extractions, and mixed-script artifacts.

\paragraph{Acceptance rule.} A sample is accepted when at least two of the three verifiers in the pool produce responses judged equivalent to the generated answer under the equivalence relation defined above, averaged across their sampled responses. Mismatched samples are discarded. We deliberately do not attempt to repair them: a disagreement is ambiguous by construction --- it may reflect a hallucinated or otherwise incorrect generated answer, or simply a question the verifier pool cannot solve. Discarding such samples is the safer policy, even at the cost of throughput.

\paragraph{Retention.} The verification stage retains approximately 35--40\% of the candidate samples. Inspection of the discarded portion shows that most disagreements trace to residual PDF text-layer artifacts that survive the validation stage (Appendix~\ref{app:text-layer}) --- specifically, broken cell understanding in tables and positional misalignment in sparse-text layouts such as forms and receipts. In both cases, the generated answer is technically supported by the source text but cannot be reliably reproduced from the rendered page, so we treat such samples as low-confidence supervision and exclude them from training.

\subsection{Text Refinement}
\label{app:text-refinement}

During evaluation after the initial post-training stage, we additionally observed that the model exhibited degraded adherence to long and detailed formatting instructions. To address these issues, we performed a refinement stage over a subset of prompts and corresponding ground-truth annotations, using Qwen3.5-397B to jointly modify the value formats in the JSON schemas and augment the prompts with detailed descriptions of the new formatting requirements. The refinement procedure consists of two stages.

\paragraph{Annotation format transformation.} Each value is rewritten into a new target format under explicit rules for common types: dates, numbers, names, URLs, and sentences. Other values undergo auxiliary normalization via abbreviations and case transformations.

\paragraph{Prompt augmentation.} The prompt is extended with matching format specifications; unmodified fields are marked ``return as in the document.'' Only specifications are added, no answers or example values, preventing supervision leakage. This recasts the task from pure extraction into joint extraction and normalization. The full prompt is in Figure~\ref{fig:refinement-prompt-step1},~\ref{fig:refinement-prompt-step2}.

\begin{figure*}[!htbp]
\centering
\begin{subfigure}{\linewidth}
\begin{minipage}{0.98\linewidth}
\begin{Verbatim}[fontsize=\small, frame=single, breaklines=true, breakanywhere=true]
You receive a (prompt, expected answer) pair extracted from a document via OCR. Perform two steps.

STEP 1: Modify the format of the expected answer.
Identify the current format of the answer, change it to a new format, and add information about the new
JSON output format to the prompt. Apply the following rules:

1. If a field in the expected answer contains only a date, identify its current format and change it
   to a new one. Example date formats: DD.MM.YYYY, DD/MM/YYYY, YYYY-MM-DD, YYYY/MM/DD, etc.
   Month-as-word formats are also allowed (day and year in digits, month spelled out); the suffix "y."
   or "year" may be added or removed after the year. For date ranges, use range formats like
   DD.MM.YYYY - DD.MM.YYYY.
2. If the answer contains a number, identify its current format and change it:
   A. Change the thousands separator (spaces or commas).
   B. If the number has a fractional part, change the decimal separator (typically comma or point),
      and either remove the trailing zero (if it does not change the value) or add an extra trailing zero.
3. If the answer contains a personal name:
   A. Identify the current grammatical case and change it to another.
   B. Identify the current name format and change it. Example formats: "Surname I.O.",
      "Surname Name Patronymic", "I.O. Surname". Important: if initials are abbreviated in the source,
      they cannot be expanded into full Name and Patronymic.
4. If the answer contains a URL:
   A. Change the URL format. Example formats: "https://www.example.com", "http://www.example.com",
      "www.example.com", "example.com".
5. If the answer contains a sentence:
   A. Change the letter case (capitalize the first letter, uppercase, lowercase, etc.).
   B. Add or change the terminal punctuation (period, exclamation mark, question mark).

General replacement rules:
1. If the answer contains words with commonly accepted abbreviations, replace them (e.g., "g." instead
   of "gorod", "ul." instead of "ulitsa").
2. If it contains abbreviations with commonly accepted full forms, expand them (e.g., "dom" instead of
   "d.", "kvartira" instead of "kv.").
3. If the answer is a single word or short phrase, identify the grammatical case of nouns and
   adjectives and change it. If the answer is a full sentence (three words or more), do not change
   the case.
4. Do not insert either the current or the modified expected answer into the modified prompt.

You may only replace existing fields with new ones; you cannot add new words without replacement, nor
remove existing words without replacement.
\end{Verbatim}
\end{minipage}
\subcaption{Step~1: annotation format transformation.}
\label{fig:refinement-prompt-step1}
\end{subfigure}
\caption{Text-refinement prompt (translated from Russian).}
\end{figure*}

\begin{figure*}[!htbp]
\ContinuedFloat
\centering
\begin{subfigure}{\linewidth}
\begin{minipage}{0.98\linewidth}
\begin{Verbatim}[fontsize=\small, frame=single, breaklines=true, breakanywhere=true]
STEP 2: Modify the current prompt to align with the modified expected answer.
Modify the prompt so that it contains information about the JSON output format, but not the answer
itself. Verify that neither the old nor the new expected answer appears in the modified prompt; if it
does, remove it. Apply the following rules:

1. If the modified answer contains a date field, add the required date format to the schema
   description (e.g., DD.MM.YYYY, YYYY-MM-DD). If the date appears inside a sentence, the format need
   not be specified. If the month is written as a word with day and year in digits, indicate this
   instead of specifying a format string.
2. If the answer contains a number, add a description of the required format:
   A. Specify whether thousands are separated and how (spaces or commas). Skip if not separated.
   B. Specify whether there is a fractional part and which separator is used. Skip if absent.
   C. Specify whether a unit or currency is required and whether a space separates them. Skip otherwise.
3. If the answer contains a personal name:
   A. Specify the required grammatical case.
   B. Specify the required format (e.g., "Surname I.O.", "Surname Name Patronymic", "I.O. Surname").
4. If the answer contains a URL:
   A. Specify the required format (https://, http://, www., or no protocol).
5. If the answer is a sentence:
   A. Specify the required letter case.
   B. Specify the required terminal punctuation.
6. If a value was not modified, state that the field should be returned in the same format as in the
   document.

General rules:
1. Clarify whether abbreviations should be used.
2. Clarify whether acronyms or full names should be used (e.g., "OOO" vs. "Limited Liability Company").
3. Specify grammatical case only when the answer is a single word or short phrase, not for full
   sentences, and never for numbers.
4. Clarify capitalization conventions.
5. Additional rules specifying formatting required for downstream parsing are allowed.
6. Do not modify already-written text unnecessarily.
7. The expected answer must not appear in the modified prompt.

Before responding, verify once more that all rules are followed and that neither the old nor the
modified answer appears in the prompt in any form. Do not list possible answer variants or examples.

DO NOT ADD ANY EXAMPLE ANSWERS OR ABBREVIATIONS TO THE PROMPT. THIS IS VERY IMPORTANT!

Response format: Return only the modified expected answer and modified prompt as JSON, without
explanations: {"modified prompt with new format description": "string",
"modified expected answer": "string"}. Do not add new fields.

Example:
Prompt: "Extract the full name of a person who is a member of the control bodies.
Response JSON format: { "name": "string" }"
Expected answer: { "name": "Ivanov Ivan Ivanovich" }
Response: {"modified prompt with new format description": "Extract the full name of a person who is a
member of the control bodies. Return the 'name' field in the instrumental case, in the format
'Surname I.O.', all letters in the field must be lowercase. Response JSON format: { 'name': 'string' }",
"modified expected answer": "{ 'name': 'ivanov i.i.' }"}
\end{Verbatim}
\end{minipage}
\subcaption{Step~2: prompt augmentation with format specifications.}
\label{fig:refinement-prompt-step2}
\end{subfigure}
\caption{Text-refinement prompt (translated from Russian), continued.}
\end{figure*}

\subsection{DADC Component Ablation}
\label{app:dadc-ablation}

To validate that each DADC stage contributes positively, we ablate the pipeline by progressively adding components on top of the internal-only baseline, holding training budget fixed (Table~\ref{tab:dadc_ablation}).

\begin{table}[!htbp]
  \centering
  \resizebox{\columnwidth}{!}{%
  \begin{tabular}{@{}lccccc@{}}
    \toprule
    \textbf{Configuration} & \textbf{SP} & \textbf{MWS} & \textbf{MP} & \textbf{FT} & \textbf{Avg w/o FT} \\
    \midrule
    Base (no SFT)                       & 0.827 & 0.677 & 0.697 & 0.722 & 0.733 \\
    + in-house                          & 0.835 & 0.685 & 0.738 & 0.951 & 0.753 \\
    \quad + open (unfiltered)           & 0.714 & 0.623 & 0.647 & 0.939 & 0.661 \\
    \quad + open (stage 1)              & 0.831 & 0.685 & 0.740 & 0.951 & 0.752 \\
    \quad + open (sampled)              & 0.833 & 0.686 & 0.747 & 0.951 & 0.755 \\
    \quad + open (verified + sampled)   & \textbf{0.842} & \textbf{0.700} & \textbf{0.764} & \textbf{0.956} & \textbf{0.768} \\
    \quad + open (verified + sampled, w/o aug)           & 0.829 & 0.681 & 0.741 & 0.950 & 0.750 \\
    \bottomrule
  \end{tabular}%
  }
  \caption{DADC pipeline ablation. All rows from row~3 onward include in-house data; ``open'' variants differ only in filtering. ``Sampled'' adds Stage~1+2 layout and fact-extractability filtering (Appendix~\ref{app:document-sampling}); ``verified'' adds cross-model consistency verification. Augmentation for open data is enabled in all experiments.}
  \label{tab:dadc_ablation}
\end{table}

Unfiltered open-domain data degrades the in-house--only baseline, as naively scraped PDFs displace production-relevant signal. Layout and fact-extractability sampling recover parity, making open data merely non-harmful; only cross-model consistency verification strictly improves over in-house--only, converting open data from a neutral additive into a positive scaling factor and indicating the gain generalizes beyond internal distributions.

\begin{table}[!htbp]
  \centering
  \small
  \begin{tabular}{lrr}
    \toprule
    \textbf{Score} & \textbf{before DADC} & \textbf{after DADC} \\
    \midrule
    0  & 0.00\%  & -  \\
    1  & 0.61\%  & -  \\
    2  & 51.68\% & - \\
    3  & 10.20\% & - \\
    4  & 16.57\% & - \\
    5  & 9.80\%  & -  \\
    6  & 8.33\%  & 69.77\%  \\
    7  & 1.93\%  & 21.32\%  \\
    8  & 0.51\%  & 6.57\%  \\
    9  & 0.37\%  & 2.32\%  \\
    10  & 0.00\%  & 0.00\%  \\
    \bottomrule
  \end{tabular}
  \caption{Difficulty distribution of the visual score of accepted samples after DADC.}
  \label{tab:dadc_ablation_visual}
\end{table}

Table~\ref{tab:dadc_ablation_visual} presents difficulty distribution after DADC pipeline. The raw distribution is heavily skewed toward easy examples. After filtering, the retained pool shifts toward higher difficulty samples while preserving the relative propotions among them.

\section{Offline Experiments}
\label{app:hyperparams}

Table~\ref{tab:hyperparams} reports the full training configuration. All runs use Supervised Fine-Tuning on Qwen3.5-35B-A3B-Base. We initially trained with full-weight updates, and later switched to LoRA adapters covering all linear layers (including those within the image encoder), all shared expert layers, and all routed expert layers.

\begin{table}[!htbp]
  \centering
  \small
  \resizebox{\columnwidth}{!}{%
  \begin{tabular}{ll}
    \toprule
    \textbf{Setting} & \textbf{Value} \\
    \midrule
    Base model            & Qwen3.5-35B-A3B-Base \\
    Trainable params      & 0.8B (LoRA) \\
    LoRA placement        & All linear layers \\
                          & including image encoder, \\
                          & shared and routed expert layers; \\
                          & router excluded \\
    Training Dataset      & 3B tokens \\
    Cutoff length         & 8{,}192 \\
    Training steps        & 5{,}120 \\
    Effective batch size  & 384 \\
    Optimizer             & AdamW \\
    Learning rate         & $2 \times 10^{-6}$ \\
    LR scheduler          & Cosine with 6\% warmup \\
    Gradient clipping     & Norm, threshold = 1 \\
    Hardware (LoRA)       & $2 \times 8$ H100 \\
    Hardware (full FT)    & $4 \times 8$ H100 \\
    \bottomrule
  \end{tabular}%
  }
  \caption{Training configuration for the main
  Supervised Fine-Tuning run.}
  \label{tab:hyperparams}
\end{table}

\subsection{LoRA vs.\ Full Fine-Tuning}
\label{app:lora-vs-full}

Our initial and final training configuration used full-parameter fine-tuning. While final downstream metrics across all benchmark groups were strong, the run converged substantially more slowly and required larger GPU FLOPs compared to a subsequent LoRA configuration we evaluated. The LoRA run reached metrics within noise of the full fine-tune at a fraction of the compute (Table~\ref{tab:lora-vs-full}), which makes it a convenient setup for rapid hypothesis testing and ablations over data mixtures, prompt formats, and curriculum choices.

\begin{table}[!htbp]
  \centering
  \small
  \resizebox{\columnwidth}{!}{%
  \begin{tabular}{@{}lcccccc@{}}
    \toprule
    \textbf{Config} & \textbf{SP} & \textbf{MWS} & \textbf{MP} & \textbf{FT} & \textbf{Avg w/o FT} \\
    \midrule
    Full FT  & 0.845 & 0.696 & 0.758 & 0.956 & 0.766 \\
    LoRA     & 0.842 & 0.700 & 0.764 & 0.956 & 0.767 \\
    \bottomrule
  \end{tabular}%
  }
  \caption{LoRA vs.\ full fine-tuning on Qwen3.5-35B-A3B-Base. Both configurations use identical data mixtures; each is reported under its own optimal hyperparameter setup, with the checkpoint selected by minimum validation loss.}
  \label{tab:lora-vs-full}
\end{table}

\subsection{Document Augmentation and Serving}
\label{app:dataserv}
\begin{figure*}[!htbp]
  \centering
  \includegraphics[width=\textwidth]{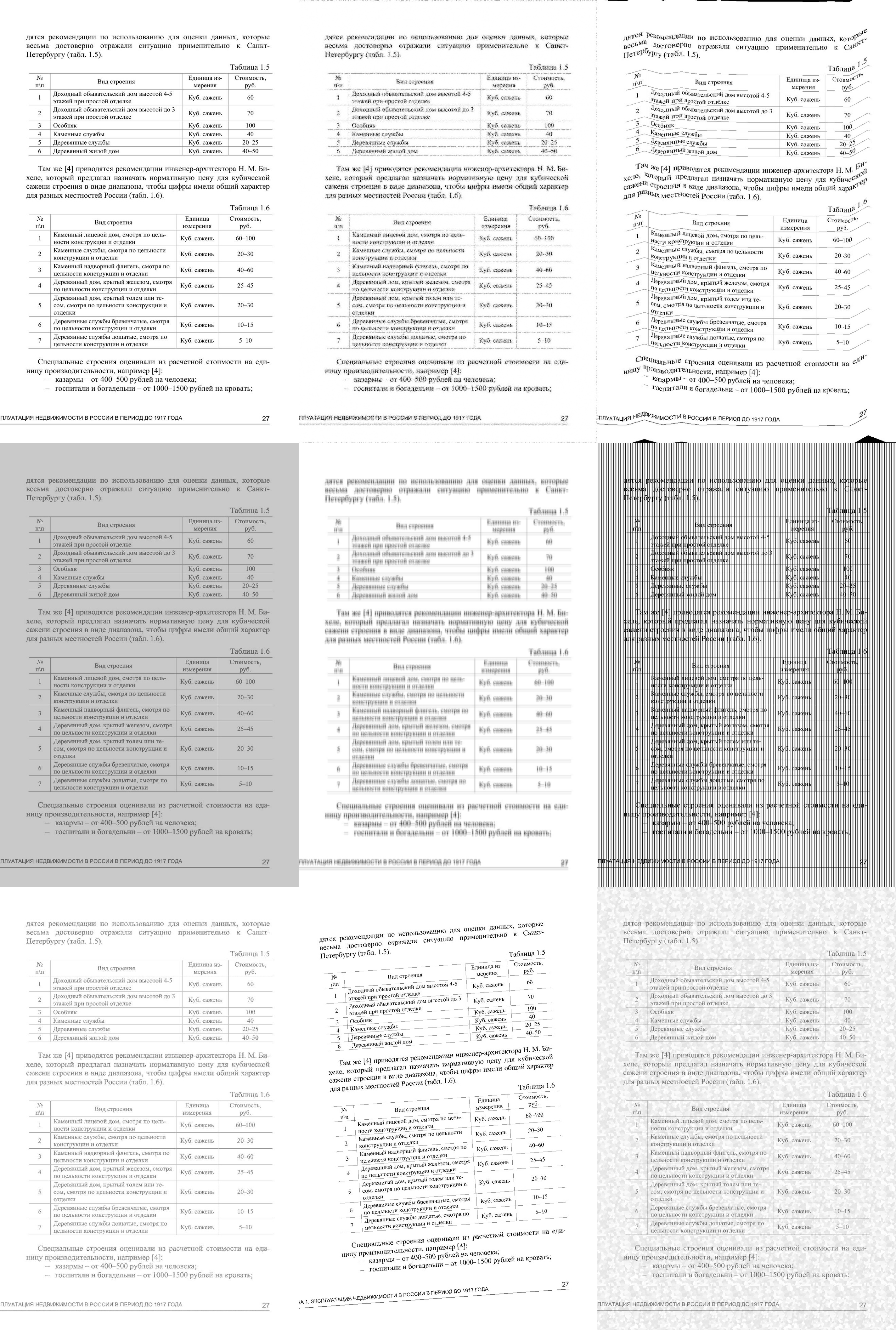}
  \caption{Examples of rendered-document augmentations applied at sample-serving time.}
  \label{fig:augmentation-grid}
\end{figure*}

External documents in our synthetic corpus are pristine white PDF pages with crisp text --- a distribution shift relative to production scans, photos, and screenshots. To close this gap, we apply rendered-document augmentations. The image-preparation stack is decoupled from training into a framework-agnostic service. The service asynchronously handles retrieval, augmentation, and pre-processing, delivering a steady stream of ready inputs to training nodes. Figure~\ref{fig:augmentation-grid} gives an example of augmented CC document. Two operational benefits motivate this architecture:

\begin{itemize}
    \item \textbf{Storage decoupling.} Images can live on storage separate from training nodes, removing the coupling between data volume and per-node disk capacity.
    \item \textbf{Dependency isolation.} Augmentations are implemented in Keras, which carries its own version-pinned dependency graph. Isolating augmentation from the training environment eliminates a common source of dependency conflicts --- an issue repeatedly noted in large-model technical reports. 
\end{itemize}

The architecture additionally enables independent horizontal scaling of the data pipeline, maximizing GPU utilization on training nodes.

\section{Latency Analysis}
\label{app:lat-analysis}

Table~\ref{tab:latency-workers} reports per-document latency under increasing concurrent worker load, measured with vLLM 0.19.1 at an average workload of 6K input tokens (including 5K visual) and up to 0.5K output tokens. Qwen2.5-VL-72B and our model are benchmarked on a single H100, while Qwen3.5-397B-A17B-FP8 requires an $8\times$H100 replica served under expert parallelism to fit. Our model sustains P95 latency well below the 10-second SLA across all tested concurrency levels (up to 7.4s at 5 workers), while both baselines exceed the SLA: Qwen2.5-VL-72B breaches the threshold already at 2 concurrent workers (P95 = 10.96s), and Qwen3.5-397B-A17B-FP8 crosses it at 3 workers (P95 = 10.77s) despite its $8\times$ larger GPU footprint, with higher concurrency levels infeasible to benchmark under the SLA constraint.

\begin{table}[!htbp]
  \centering
  \footnotesize
  \setlength{\tabcolsep}{3pt}
  \begin{tabular}{@{}ccccccc@{}}
    \toprule
    \multirow{2}{*}{\textbf{W}}
      & \multicolumn{2}{c}{\textbf{Qwen2.5-72B}}
      & \multicolumn{2}{c}{\textbf{Ours}}
      & \multicolumn{2}{c}{\textbf{Qwen3.5-397B}} \\
    \cmidrule(lr){2-3} \cmidrule(lr){4-5} \cmidrule(lr){6-7}
      & P50 & P95 & P50 & P95 & P50 & P95 \\
    \midrule
    1 & \textbf{6.89}  & \textbf{7.26}  & \textbf{5.65} & \textbf{5.74} & \textbf{8.90}  & \textbf{8.96}  \\
    2 & 10.25 & 10.96 & \textbf{6.13} & \textbf{6.26} & \textbf{9.63}  & \textbf{9.72}  \\
    3 & 13.58 & 14.35 & \textbf{6.76} & \textbf{6.90} & 10.67 & 10.77   \\
    4 & 16.86 & 17.73 & \textbf{7.17} & \textbf{7.36} & 11.16   & 11.73 \\
    5 & 20.21 & 23.86 & \textbf{7.19} & \textbf{7.41} & 12.79   & 13.30 \\
    \bottomrule
  \end{tabular}
  \caption{Per-document latency (P50 / P95, seconds) under increasing concurrent worker load $W$.}
  \label{tab:latency-workers}
\end{table}

\section{Cost Model Details}

\subsection{Calibration of Assisted-Mode Coefficients}
\label{app:cost-coeff-calibration}
The constants $0.3x$ (confirmation) and $1.1x$ (correction) are are derived from a retrospective case study on a pool of 50 trained annotators working with already-deployed document-processing ML models in production.

\paragraph{Setup.} We selected 50 trained annotators on our production document labeling team. For each annotator we extracted from production telemetry every field-level event over a rolling three-month window, yielding several million field interactions in total. Each event was tagged with the operator's mode --- \emph{manual}, \emph{confirm}, or \emph{correct} and per-field handling latency. The manual rate was estimated as the median per-field latency across all manual-mode events, restricted to fields where no ML suggestion was available. 

\paragraph{Confirmation coefficient.} The confirmation coefficient was estimated as the median per-field latency in confirm-mode events, divided by the manual rate. Across the 50-annotator cohort the per-annotator estimates clustered tightly in $[0.27, 0.34]$, with cohort-level median $0.30$. Stability across annotators reflects the fact that read-and-confirm is bottlenecked by visual verification time, which varies less across operators than search-and-transcribe.

\paragraph{Correction coefficient.} The correction coefficient was estimated analogously on correct-mode events, where the operator overrode a pre-filled value. The cohort-level median was $1.08$, with per-annotator estimates in $[1.04, 1.17]$. The overhead above $1.0$ isolates the cost of error detection and the context-switching penalty of discarding a displayed suggestion. We round to $1.1$ in the main paper to retain a single significant figure consistent with the confirmation coefficient.

\end{document}